\documentclass[11pt,a4paper]{article}

\usepackage[utf8]{inputenc}
\usepackage[T1]{fontenc}
\usepackage{amsmath,amssymb,amsthm}
\usepackage{mathtools}
\usepackage{bm}
\usepackage{graphicx}
\usepackage{booktabs}
\usepackage{array}
\usepackage{longtable}
\usepackage{hyperref}
\usepackage[margin=1in]{geometry}
\usepackage{enumitem}
\usepackage{xcolor}
\usepackage{natbib}
\usepackage[expansion=false]{microtype}

\theoremstyle{definition}
\newtheorem{definition}{Definition}
\newtheorem{axiom}{Axiom}
\newtheorem{proposition}{Proposition}
\newtheorem{corollary}{Corollary}
\newtheorem{remark}{Remark}

\newcommand{\Esub}{\ensuremath{E_3}}
\newcommand{\Lrec}{\ensuremath{L_{\mathrm{rec}}}}
\newcommand{\Cstruct}{\ensuremath{C_{\mathrm{struct}}}}
\newcommand{\Cupdate}{\ensuremath{C_{\mathrm{update}}}}
\newcommand{\Cremain}{\ensuremath{C_{\mathrm{remain}}}}
\newcommand{\Ctot}{\ensuremath{C_{\mathrm{tot}}}}
\newcommand{\Ssys}{\ensuremath{\Sigma_{\mathrm{sys}}}}
\newcommand{\Sanchor}{\ensuremath{\Sigma_{\mathrm{anchor}}}}
\newcommand{\Ssense}{\ensuremath{\Sigma_{\mathrm{sense}}}}
\newcommand{\Sapriori}{\ensuremath{\Sigma_{\mathrm{apriori}}}}
\newcommand{\St}{\ensuremath{\Sigma_t}}
\newcommand{\At}{\ensuremath{A_t}}
\newcommand{\Pt}{\ensuremath{\Pi_t}}
\newcommand{\CDt}{\ensuremath{\mathrm{CD}_t}}
\newcommand{\Ht}{\ensuremath{H_t}}
\newcommand{\Bt}{\ensuremath{B_t}}

\newcommand{\costar}{\ensuremath{\mathrm{co}^*}}
\newcommand{\simt}{\ensuremath{\mathrm{sim}_t}}
\newcommand{\retr}{\ensuremath{\mathrm{retr}_t}}
\newcommand{\replay}{\ensuremath{\mathrm{replay}}}
\newcommand{\emb}{\ensuremath{\mathrm{emb}}}
\newcommand{\cU}{\ensuremath{\mathcal{U}}}
\newcommand{\cE}{\ensuremath{\mathcal{E}}}
\newcommand{\cP}{\ensuremath{\mathcal{P}}}

\title{\textbf{An Axiomatic Approach to General Intelligence: SANC(\texorpdfstring{$E_3$}{E3})}\\[0.5em]
\large Self-organizing Active Network of Concepts with Energy $E_3$}

\author{
Daesuk Kwon$^{1}$ \and Won-gi Paeng$^{1}$\\[0.5em]
$^{1}$Hyntel, Inc., Seoul, Republic of Korea\\
\texttt{hyntel@hyntel.com}, \texttt{wgpaeng@hyntel.net}
}

\date{}

\begin{document}

\maketitle

\begin{abstract}
General intelligence must reorganize experience into internal structures that enable prediction and action under finite resources. Existing systems implicitly presuppose fixed primitive units---tokens, subwords, pixels, or predefined sensor channels---thereby bypassing the question of how representational units themselves emerge and stabilize.

This paper proposes Self-organizing Active Network of Concepts with Energy $E_3$ -- \textbf{SANC($E_3$)}, an axiomatic framework in which representational units are not given \emph{a priori} but instead arise as stable outcomes of competitive selection, reconstruction, and compression under finite activation capacity, governed by the explicit minimization of an energy functional $E_3 = \lambda_1 L_{\mathrm{rec}} + \lambda_2 C_{\mathrm{struct}} + \lambda_3 C_{\mathrm{update}}$.

SANC($E_3$) draws a principled distinction between \textbf{system tokens}---structural anchors such as \{here, now, I\} and sensory sources---and tokens that emerge through self-organization during co-occurring events. Five core axioms formalize finite capacity, association from co-occurrence, similarity-based competition, confidence-based stabilization, and the reconstruction--compression--update trade-off. Residual capacity $C_{\mathrm{remain}}(t)$ is introduced as an explicit control variable that dynamically modulates representation creation, stabilization, and deletion thresholds.

A key feature of the framework is a \textbf{pseudo-memory-mapped I/O mechanism}, through which internally replayed Gestalts are processed via the same axiomatic pathway as external sensory input. As a result, perception, imagination, prediction, planning, and action are unified within a single representational and energetic process. This unification naturally extends the framework to embodied and physical agents, where interaction with the environment and motor behavior are treated as continuations of Gestalt completion rather than as separate control modules.

From the axioms, twelve propositions are derived, showing that category formation, automatic threshold tuning, hierarchical organization, unsupervised learning, and high-level cognitive activities---dialogue, authoring, perception, causal inference, and action---can all be understood as instances of \textbf{Gestalt completion under $E_3$ minimization}.

\end{abstract}

\section{Introduction}

For millennia, the question ``What is intelligence?'' has been posed in myriad forms. Contemporary artificial intelligence has inherited this question yet still lacks a rigorous, operational definition of general intelligence. Despite impressive recent advances---most notably attention-based Transformers~\cite{vaswani2017attention}---humanity has arguably not achieved AGI. We contend that this shortfall stems from attempts to implement isolated facets of intelligence on the basis of accumulated empirical success and local mechanistic insight, without first defining what intelligence is, or worse, by adopting an inadequate definition. In the absence of a definition, one cannot deliberately engineer general intelligence; at best, one may stumble toward something that resembles it.

Rather than attempting a direct definition of `general intelligence,' we take human intelligence as the paradigmatic case: whatever humans do when we say they are `being intelligent'---that is what we mean by general intelligence. Our approach is to ask: what structural and dynamical properties must a system possess to exhibit such intelligence? The axioms we propose are intended as necessary conditions that human intelligence appears to satisfy. Any system satisfying all these axioms is termed \emph{SANC($E_3$)-intelligent}. To our knowledge, human intelligence remains the only known system that fully satisfies these axioms; whether artificial systems can do so is precisely the question this framework aims to clarify.

This paper proposes \textbf{SANC($E_3$)} (\emph{Self-organizing Active Network of Concepts with Energy $E_3$}), a minimal axiomatic framework for characterizing the emergence of intelligence.

SANC($E_3$) distinguishes \emph{system tokens} (given by the architecture) from \emph{tokens} that emerge through self-organization. It defines an intelligent system as one that, under finite resources and an explicit energy-minimization objective, autonomously generates and organizes representational units whose experiential completion yields prediction and can trigger action. Such representational units are termed Gestalts.

In SANC($E_3$), association---not binding---is taken as the primitive of intelligence: intelligence is defined as the capacity to map co-occurring events to associated internal representations under finite resources, while compositional operators such as $G$ are treated solely as formal realizations of this associative relation.

We do not claim a complete axiomatization in the strict sense of mathematical logic; rather, we present a principled axiomatic framework specifying necessary conditions for general intelligence---understood here as the capacity to autonomously form, organize, and deploy internal representations that enable prediction and action across open-ended domains under finite resources.

A crucial feature of the framework is that characteristics such as morpheme-level tokens and unsupervised learning are not externally imposed design goals but natural consequences of the axiomatic system. Likewise, causal inference and social interaction are explained naturally by the same Gestalt-completion principle, without dedicated specialized mechanisms.

\subsection{Related Work}

\textbf{Formal definitions of intelligence.} Efforts to define intelligence mathematically or axiomatically have been central to AGI research. Hutter's AIXI model~\cite{hutter2005universal} defines an optimal agent by combining Solomonoff induction with sequential decision theory, maximizing expected reward while favoring simpler environment models. AIXI is, however, incomputable and presupposes infinite resources; moreover, it does not address how representational units (tokens) form. Legg and Hutter~\cite{legg2007universal} extend this idea to ``universal intelligence,'' defined as a weighted sum of goal achievement across environments, but the same limitations persist.

Wang's NARS~\cite{wang2019defining} defines intelligence as adaptation under insufficient knowledge and resources (AIKR). SANC's A1 (finite capacity) aligns with AIKR's resource constraints, yet SANC goes further: it explains how representational units themselves emerge (T3) and how hierarchical structure forms (T7, T8) under an explicit energy decomposition.

Chollet~\cite{chollet2019measure} characterizes intelligence as skill-acquisition efficiency and proposes an algorithmic-information-theoretic formalization; the ARC benchmark targets measurement. SANC is not primarily a measurement framework but a generative one: it specifies, via axioms, what an intelligent system must satisfy and how its core properties arise.

Friston's Free Energy Principle~\cite{friston2010free} posits that agents minimize variational free energy, emphasizing prediction-error minimization. SANC likewise adopts energy-based dynamics but decomposes $E_3$ explicitly into three terms ($\Lrec$, $\Cstruct$, $\Cupdate$) to formalize the trade-off among reconstruction accuracy, structural simplicity, and update stability.

Bostick~\cite{bostick2024agi} proposes an axiomatic definition of AGI as ``Lawful Autonomous Continuity,'' emphasizing identity continuity and autonomy. As a purely definitional approach, it does not specify mechanisms for representation formation or learning; SANC renders these mechanisms explicit through axioms and derived propositions.

\textbf{Distinctive features of SANC($E_3$).} Compared to prior work, SANC($E_3$) is distinguished by: (i) tokens are emergent outcomes of $E_3$ dynamics, not primitives (T3); (ii) system tokens $\Ssys$ formalize Kantian \emph{a priori} anchors to resolve bootstrapping; (iii) the trade-off between token explosion and sequence explosion is made explicit (T4); (iv) unsupervised learning is a natural consequence of the axiomatic system (T10); and (v) dialogue, authoring, causal inference, perception, and self-improvement (extended definition D15) are unified as Gestalt completion.

\textbf{Connections to recent AI research.} Forward--Forward learning~\cite{hinton2022forward} demonstrates local learning without backpropagation; SANC's A2 (similarity-based competition) and A3 (confidence-based stabilization) provide an axiomatic account of local competitive dynamics. JEPA and I-JEPA~\cite{lecun2022path,assran2023ijepa} propose predictive representation learning in embedding space; SANC's T6 generalizes this insight: \emph{Gestalt completion (replay) is prediction}. Nested Learning~\cite{behrouz2025nested} emphasizes multiple time scales; SANC captures this via A3 and self-similar levels (T8). Large Concept Models~\cite{meta2024lcm} aim to operate at concept level; SANC is more fundamental, treating both vocabulary-level tokens and concept-level tokens as outcomes generated identically by the axioms (T3). Finally, fixed tokenization schemes such as BPE~\cite{sennrich2016bpe} assume a static vocabulary; SANC requires the token inventory itself to evolve with accumulated experience (T3, T4).

\textbf{What we contribute.} (i) Core axioms A1--A5. (ii) Mechanism axioms A6--A10. (iii) Derived propositions T1--T12. (iv) Extended definitions D10--D15. (v) Systematic comparison with existing theories. (vi) Appendix~C provides the original postulates, first formulated in the mid-1990s, in informal language for intuitive accessibility.

\section{Notation}

\textbf{Terminological convention (levels).} The core theoretical entities in SANC are the \emph{associated event} and its corresponding \emph{associated representation}; an associated event is said to form a \emph{Gestalt} when it constitutes a coherent, unified whole that is distinguished and mapped to an internal associated representation. The term \emph{token} is used only as an implementation convenience (e.g., Transformer symbols used to instantiate representations). Wherever possible, we prefer the theoretical terms \emph{associated event / associated representation / Gestalt}.

We introduce a formal vocabulary. The notation is implementation-agnostic: it constrains what must exist without prescribing whether the realization is neural, symbolic, or hybrid.

\begin{itemize}[leftmargin=2em]
\item $\cE$: set of events (external stimuli or internal activations).
\item $\cE^*$: set of finite event sequences (Kleene star); $e_{1:t} \in \cE^*$ denotes the sequence from time 1 to $t$.
\item $\cU$: universe of representational units (discrete symbols, vectors, subspaces, distributions, etc.).
\item $\At \subseteq \cU$: active state at time $t$.
\item $\kappa, C$: capacity measure and finite upper bound; $\kappa(\At) \le C < \infty$.
\item $\Ctot$: total available activation capacity.
\item $\Cremain(t)$: residual capacity, defined as $\Ctot - \kappa(\At)$.
\item $\Lrec$: reconstruction loss.
\item $\Cstruct$: structural complexity cost.
\item $\Cupdate$: update cost.
\item $\cP(X)$: power set of $X$.
\item $\cot(x, y)$: predicate indicating that $x$ and $y$ are co-detected or co-active at time $t$.
\item $\costar(x, y)$: accumulated co-occurrence statistic over time; the specific increment/decrement scheme is implementation-dependent, but to maintain finite capacity, items not reactivated for an extended period may be weakened or removed.
\item $\Pt$: selection operator (the concrete rule may vary but must satisfy A2).
\item $S_t \subseteq \cU$: candidate set at time $t$ (active units, association candidates, retrieval candidates, etc.).
\item $\Pt(S_t) \subseteq S_t$: units actually selected (or reinforced) from $S_t$.
\item $\CDt(u)$: confidence degree of unit $u$ at time $t$.
\item $G$: association (compositional) operator, $G: \cU \times \cU \to \cU$.
\item $\simt$: similarity predicate on elements of $\cU$.
\item $[u]_{\mathrm{sim}}$: similarity neighborhood of $u$ (not assumed to be an equivalence class): $[u]_{\mathrm{sim}} := \{ v \in S_t : \simt(u, v) \}$.
\item $\Esub$: energy function, weighted sum of $\Lrec$, $\Cstruct$, $\Cupdate$ (Axiom A4).
\item $\theta_{\mathrm{stab}}$: stabilization threshold (token-formation threshold).
\item $\theta_{\mathrm{err}}$: reconstruction-error threshold (reconstruction succeeds iff $\Lrec \le \theta_{\mathrm{err}}$).
\item $P_t(\cdot)$: predictive distribution over the next event at time $t$ (e.g., a normalized score distribution).
\item $\Ht$: predictive uncertainty (e.g., entropy $\Ht = -\sum P_t(e) \log P_t(e)$, sum over $e \in \cE$).
\item $\theta_{\mathrm{unc}}$: uncertainty threshold (context segmentation and ``completed Gestalt candidate'' detection).
\item $\Bt$: the current associated event (Gestalt) formed up to time $t$ (A8).
\item $\mathrm{terminate}(\Bt)$: operator that finalizes $\Bt$ as a ``completed Gestalt candidate'' and ends the current association (A8).
\item $\mathrm{reset}_{\mathrm{assoc}}(t+1)$: operator that resets the association state so a new Gestalt association begins at time $t+1$ (A8).
\item $S_i$: the $i$-th SANC($E_3$)-intelligent subsystem within a larger system (A10).
\item $\mathrm{Comm}(i, j)$: a communication relation/channel enabling Gestalt exchange between subsystems $S_i$ and $S_j$ (A10).
\item $\mathrm{send}_{i \to j}(\mathfrak{g})$: transfer operator that sends a Gestalt (or a Gestalt sequence/set) $\mathfrak{g}$ from $S_i$ to $S_j$ (A10).
\item $\mathrm{recv}_{j \leftarrow i}(\mathfrak{g})$: reception operator by which $S_j$ receives $\mathfrak{g}$ from $S_i$; received Gestalts are elements of the shared universe $\cU$ and are included in the candidate set $S_t$, participating in A5--A8 (A10).
\item $\St$: set of stabilized tokens at time $t$, $\St := \{ u \in \cU : \CDt(u) \ge \theta_{\mathrm{stab}} \}$.
\item $\tau$: a token at time $t$, $\tau \in \St$.
\item $\Ssys$: system tokens: $\Sanchor \cup \Ssense \cup \Sapriori$ (see D1).
\item $\Sanchor$: anchor tokens (e.g., \{here, now, I\}; see D1).
\item $\Ssense$: sensory tokens.
\item $\Sapriori$: (\emph{a priori}) intrinsic tokens.
\end{itemize}

\textbf{Auxiliary notation for selection bias:}
\begin{itemize}[leftmargin=2em]
\item $\mathrm{score}_t(u)$: scalar selection score for candidate $u$ (implementation-defined).
\item $\Delta_t^{\mathrm{will}}(u)$: will/intent bias term added to the score.
\item $\mathrm{score}_t^{\mathrm{will}}(u) := \mathrm{score}_t(u) + \Delta_t^{\mathrm{will}}(u)$.
\item $\Pi_t^{\mathrm{will}}(S_t)$: will-biased selection (see D11).
\item $\delta_I$: I-proximity distance threshold (see D12).
\end{itemize}

\begin{remark}[distributed vs.\ discrete]
Elements of $\cU$ need not be discrete symbols. A ``token'' is a stabilized unit in $\Sigma$---an outcome of stabilization, not an \emph{a priori} primitive. (\textbf{Only system tokens} are tokens in $\St$ that are given \emph{a priori}.)
\end{remark}

\begin{remark}[homogeneity across levels]
In this paper, `homogeneity across levels' means that the axiomatic rules governing event--representation correspondence, association/generation, selection/retention (and deletion of non-\emph{a priori} representations), and finite capacity constraints operate identically regardless of level.
\end{remark}

\section{Core Definitions}

\begin{definition}[D0: SANC($E_3$)-intelligence]
A system is SANC($E_3$)-intelligent if there exist a representational universe $\cU$, an activation dynamic $\At$ over $\cU$, a capacity measure $\kappa$ with finite upper bound $C$, a candidate set $S_t$, a selection operator $\Pt$, a confidence degree $\CDt$, system tokens $\Ssys$, and an energy $\Esub = \lambda_1 \Lrec + \lambda_2 \Cstruct + \lambda_3 \Cupdate$ such that Axioms A1--A10 hold and the system's dynamic update mechanism operates to minimize $\Esub$ (at least approximately) under the constraints of A1--A3---specifically, the update dynamics exhibit a descent property that decreases $\Esub$ over time (in expectation).
\end{definition}

\begin{definition}[D1: representational unit, identification, segmentation, system tokens]
A representational unit is any $u \in \cU$. An identification operator $\mathrm{id}: \cE \to \cU$ assigns a representational unit to an event. A segmentation operator $\mathrm{seg}: \cU \to \cP(\cU)$ decomposes a unit into parts. System tokens $\Ssys \subseteq \cU$ are representational units given by the system's structure, rather than emerging from self-organization through event association.

\begin{itemize}
\item \textbf{Anchor tokens $\Sanchor$:} structurally given tokens such as \{here, now, I\} that co-occur with, and are embedded with, all representations.
\item \textbf{Sensory-source tokens:} outputs of sensory receptors---level-0 materials.
\end{itemize}

Anchor tokens participate in the same network as other tokens, serving as relational reference points, but their existence is presupposed. Their interpretation (what ``here'' or ``now'' refers to) may evolve through experience.

\emph{Relation to Kant.} System tokens can be viewed as formalizing Kantian \emph{a priori} forms that structure experience. Unlike Kant's forms~\cite{kant1781kritik}, however, system tokens are themselves tokens: they participate as nodes in the relational network.
\end{definition}

\begin{definition}[D2: distinction and consistency]
The system can distinguish identical, similar, and different events, and can assign consistent internal representations to them. Specifically: identical events map to the same representation; similar events decompose into shared and differing parts (see D3).
\end{definition}

\begin{definition}[D3: similarity]
$\simt(u, v) \iff \exists\, a, b, c: u = G(a, b), v = G(a, c), b \ne c$. Here $a$ is the shared subcomponent; $b$ and $c$ are the differing parts. Thus similarity requires (i) at least one shared component, (ii) at least one differing component, and (iii) recomposability into $u$ and $v$.

\emph{Example.} Suppose $ab = G(a, b)$ corresponds to ``with Mom'' and $ac = G(a, c)$ to ``with Dad.'' Then $\simt(ab, ac)$ holds: $a =$ ``with'' is the shared part; $b =$ ``Mom'' and $c =$ ``Dad'' are the differing parts.
\end{definition}

\begin{definition}[D4: association and associated representation]
Association is the fundamental relation by which co-occurring events are related. Formally, association may be realized by a compositional operator $G : \cU \times \cU \to \cU$, which constructs an associated representation from existing representations. The statement ``$x$ is associated with $y$'' can be understood as: $\cot(x, y)$ or $\costar(x, y)$ causes $G(x, y)$ to become a candidate for selection. An association (and its realized associated representation) can be strengthened or weakened by selection $\Pt$ and by the frequency of observation (experience), and, unlike \emph{a priori} representations, is not permanent. $\costar(x, y)$ denotes the temporal accumulation of co-occurrence, increasing or decreasing with co-occurrence frequency; the specific increment/decrement scheme is implementation-dependent.
\end{definition}

\begin{definition}[D5: Gestalt]
A Gestalt is an associated structure that (i) relates co-occurring components, (ii) can become a selectable unit under $\Pt$, (iii) is internally replayable, and (iv) can enter $\St$ upon stabilization. It is a representational unit in formation or already formed.

This formalizes the Gestalt-psychology dictum (Wertheimer~\cite{wertheimer1923laws}) that ``the whole is more than the sum of its parts.'' Once formed, a Gestalt can be treated as an event at a higher level and can be associated to others again (see A6).
\end{definition}

\begin{definition}[D6: stabilized set and tokens]
$\St := \{ u \in \cU : \CDt(u) \ge \theta_{\mathrm{stab}} \}$. A token at time $t$ is any $\tau \in \St$. (System tokens are presupposed separately.)

\emph{Remark.} Tokens other than system tokens are understood to form when Gestalt candidates activated by experience undergo association, selection, and reinforcement and stabilize into $\St$. Moreover, thresholds such as $\theta_{\mathrm{stab}}$ are not assumed to be fixed constants but \textbf{are adaptively set to avoid increasing $\Esub$} so as not to worsen the internal objective $\Esub$ (see Proposition T12).
\end{definition}

\begin{definition}[D7: category]
A \emph{category} is a token that captures shared parts or context across multiple representational units and thereby lowers $\Esub$.

\emph{Example.} Having repeatedly experienced $ab =$ ``with Mom'' and $ac =$ ``with Dad,'' followed by $db =$ ``for Mom'' and $dc =$ ``for Dad,'' two new categories \{with, for\} and \{Mom, Dad\} may emerge---corresponding, for instance, to preposition and noun categories.
\end{definition}

\begin{definition}[D8: retrieval, replay/regeneration, and processing equivalence]\leavevmode
\begin{enumerate}[label=(\alph*)]
\item \textbf{Partial-match retrieval.} Partial-match retrieval is an operator $\retr: \cE^* \to \cP(S_t)$ that, given an observed event history $e_{1:t}$, returns a set of candidate Gestalts in $S_t$ sharing matching parts with the current evidence. That is, $\retr(e_{1:t}) \subseteq S_t$, and this set may contain stabilized tokens ($\St$) as well as pre-stabilized Gestalt candidates (not yet in $\St$). (Analogous to spreading activation in semantic networks (Quillian~\cite{quillian1968semantic}) and to Gestalt completion.)

\item \textbf{Replay/regeneration.} $\replay(u)$ denotes an internal operation that reactivates or regenerates a Gestalt $u$ so as to reduce $\Lrec$ under A4, typically via retr. Replay is a mechanism for lowering $\Lrec$ and, more broadly, part of the system's $\Esub$-minimizing dynamics.

\item \textbf{Pseudo-memory-mapped I/O.} Internal replay of a Gestalt is subject to the same axiomatic decision rules as external sensory input---namely similarity-based competition (A2), selection and stabilization (A3), evaluation of reconstruction loss $\Lrec$, and global minimization of $\Esub$ (A4).

As a result, internally replayed Gestalts drive downstream systems---including reasoning, inference, dialogue, and action generation---in a manner indistinguishable from externally perceived events. The equivalence concerns not the origin of the input but the identity of the axiomatic processing path and its causal effects.

This organization is functionally analogous to memory-mapped I/O, but is not based on physical address mapping; rather, it relies on isomorphism of representational processing rules. We therefore refer to this mechanism as pseudo-memory-mapped I/O.
\end{enumerate}
\end{definition}

\begin{definition}[D9: embedding / relational distance vector and learning]\leavevmode
\begin{enumerate}[label=(\alph*)]
\item For a representation $u \in \cU$, define an \textbf{embedding} as:
\[
\emb(u) = \bigl(\{ d(u, v) : v \in \St,\, \costar(u, v) > 0 \},\, \{ d(u, \tau) : \tau \in \Sanchor \}\bigr),
\]
where $\St$ is the stabilized token set at time $t$, $\Sanchor = \{\text{here}, \text{now}, I\} \subseteq \Ssys$ are \emph{a priori} coordinate-like tokens, and $d(u, v)$ is a relational (latent) distance. In this paper, the \textbf{`meaning'} of a token $u$ is the \textbf{relational profile} given by its relationships (similarity/distance/association) with other tokens at the same time.

Because anchor tokens are assumed to co-occur with all representations, every embedding contains distances to $\Sanchor$. Tokens such as \emph{past}, \emph{future}, or \emph{world} are not \emph{a priori} tokens but can be interpreted as tokens emerging from relational structure (such as distance to $\Sanchor$).

\emph{Remark 1:} $\emb(u)$ denotes an \textbf{(unordered) profile} of the above distances; in implementation, this may be arranged into a fixed-order/dimension vector or summarized as statistics.

\emph{Remark 2 (Saussure and relational meaning).} Defining meaning as a relational profile is consistent with Saussure's insight~\cite{saussure1916cours} that the meaning of a sign is not a fixed entity but is determined by its relations (context) with other signs (arbitrariness of the sign / system of differences). Definition D9 formalizes this as distance-based embedding.

\item \textbf{Learning.} In this paper, `learning' refers to the process by which Gestalts (tokens) in $\St$ are created, deleted, or stabilized as experience accumulates, and their relationships (similarity/association/confidence degree/inter-token relational profiles) are updated so as to better satisfy the internal objective $\Esub$.
\end{enumerate}
\end{definition}

\section{Core Axioms}

The first five axioms establish foundational constraints; the remaining five (Section~5) specify compositional and structural mechanisms. Together, they force the emergence of tokens, categories, and hierarchy without hand-designed vocabulary or rules.

\begin{axiom}[A1: finite active capacity]
$\kappa(\At) \le C < \infty$. The active state has finite capacity at every time $t$. Finiteness necessitates competition.
\end{axiom}

\begin{axiom}[A2: similarity-based competition and single selection]
$|\Pt(S_t) \cap [u]_{\mathrm{sim}}| \le 1$, and $\kappa(\Pt(S_t)) \le C$. Similar candidates compete, and at most one is selected from each similarity neighborhood.
\end{axiom}

\begin{axiom}[A3: stabilization and deletion by identification and selection]\leavevmode
\begin{enumerate}[label=(\alph*)]
\item \textbf{Stabilization by confidence degree.}
\begin{itemize}
\item If $u \in \Pt(S_t)$: $\mathrm{CD}_{t+1}(u) > \CDt(u)$ \quad [selection reinforces]
\item If $u \notin \Pt(S_t)$ repeatedly: $\CDt(u) \to 0$ \quad [non-selection extinguishes]
\end{itemize}
The specific increment/decrement scheme is implementation-dependent, subject to A4.

\item \textbf{Forgetting and deletion.} Non-\emph{a priori} representations---including those derived from experienced events or from \emph{a priori} representations---may have their $\costar$ values decrease if the corresponding event (or equivalent reactivation evidence) is not observed for a certain period, eventually leading to deletion from $\St$. (The decay scheme is left to implementation.)

\item \textbf{\emph{A priori} representation exception.} \emph{A priori} representations are exempt from the forgetting/deletion rules above.
\end{enumerate}
\end{axiom}

\begin{axiom}[A4: energy $E_3$ and reconstruction--update dynamics]\leavevmode
\begin{enumerate}[label=(\alph*)]
\item \textbf{Definition.} $\Esub = \lambda_1 \Lrec + \lambda_2 \Cstruct + \lambda_3 \Cupdate$, where:
\begin{itemize}
\item $\Lrec$ (reconstruction loss): error in reconstructing events using the current token set $\St$.
\item $\Cstruct$ (structural cost): complexity of the token inventory (e.g., number of tokens, total description length).
\item $\Cupdate$ (update cost): instability incurred by changing tokens (e.g., invalidating existing links).
\end{itemize}

\item \textbf{Intelligence condition.} A SANC($E_3$)-intelligent system acts to minimize $\Esub$.

\item \textbf{Trade-off.} The three terms constituting $\Esub$ trade off against each other as follows, and $\Esub$ minimization finds a balance:
\begin{itemize}
\item Minimizing only $\Lrec$ $\to$ token explosion ($\Cstruct \uparrow$)
\item Minimizing only $\Cstruct$ $\to$ reconstruction failure ($\Lrec \uparrow$)
\item Minimizing only $\Cupdate$ $\to$ inability to adapt ($\Lrec \uparrow$ on new experience)
\end{itemize}

\item \textbf{Reconstruction--update dynamics.} For each event $e$:
\begin{itemize}
\item The system attempts reconstruction using current $\St$.
\item If $\Lrec(e, \St) \le \theta_{\mathrm{err}}$ (success): the CD of the Gestalts used is reinforced.
\item If $\Lrec(e, \St) > \theta_{\mathrm{err}}$ (failure): association (A5), splitting (A7), and competition (A2) are triggered, updating $\St \to \Sigma_{t+1}$.
\end{itemize}

\textbf{Relation to backpropagation.} The error-driven update in (d) is compatible with error backpropagation (EBP) as an implementation mechanism. However, an EBP-only system that optimizes only reconstruction loss $\Lrec$ ignores $\Cstruct$ and $\Cupdate$, and therefore need not minimize $\Esub$ as a whole. SANC does not preclude EBP; it requires that learning operate under A1--A3 and the full $\Esub$ trade-off.

\item \textbf{Capacity--Energy Coupling.} In a SANC($E_3$)-intelligent system, the dynamics of $\Esub$ minimization are modulated by the residual capacity $\Cremain(t)$.

The thresholds associated with representation creation, stabilization, and deletion ($\theta_{\mathrm{create}}$, $\theta_{\mathrm{stab}}$, $\theta_{\mathrm{delete}}$) are not fixed constants but are dynamically adjusted as functions of $\Cremain(t)$.

As $\Cremain(t)$ decreases, the system suppresses the creation of new Gestalts and favors reuse and compression of existing ones, preventing uncontrolled growth of $\Cstruct$ while minimizing degradation of $\Lrec$.
\end{enumerate}
\end{axiom}

\begin{axiom}[A5: co-occurrence generates association candidates]
$\cot(x, y) \lor \costar(x, y) \Rightarrow \exists\, t' \ge t : G(x, y) \in S_{t'}$. Co-detected events become candidates for associated representation (Gestalt). This axiom formalizes Hebb's principle~\cite{hebb1949organization} that ``neurons that fire together wire together.'' (Association candidates thus generated may be weakened or deleted by A3 if not sufficiently reinforced/reactivated.)
\end{axiom}

\section{Mechanism Axioms}

\begin{axiom}[A6: closure under association]
For all $x, y \in \cU$, $G(x, y) \in \cU$. The universe $\cU$ is closed under association. Consequently, recursive composition and hierarchy become possible.
\end{axiom}

\begin{axiom}[A7: similarity-based splitting]
If $\simt(G(a, b), G(a, c))$ holds, a splitting occurs that identifies the shared part $a$ and the differing parts $b$ and $c$.
\end{axiom}

\begin{axiom}[A8: uncertainty-gated segmentation and Gestalt completion]
If the uncertainty $\Ht$ of the predictive distribution $P(e_{t+1} \mid e_{1:t})$ for the next event exceeds threshold $\theta_{\mathrm{unc}}$, the current associated event (Gestalt) $\Bt$ is deemed complete and association restarts at $t+1$.
\[
\Ht \ge \theta_{\mathrm{unc}} \Rightarrow \mathrm{terminate}(\Bt) \land \mathrm{reset}_{\mathrm{assoc}}(t+1).
\]

\emph{Remark.} The system may compute a predictive distribution $P_t(\cdot)$ for the next event given the observed history $e_{1:t}$, and its uncertainty $\Ht$ (e.g., entropy). As $\Ht$ decreases (predictions become more certain), the current context tends to cohere into a single Gestalt. Conversely, when a context (Gestalt) is completed and a new context begins, $\Ht$ may spike; if $\Ht \ge \theta_{\mathrm{unc}}$, the system treats the current history $e_{1:t}$ as a ``completed Gestalt candidate,'' includes it among association/stabilization candidates, and segments subsequent events as a `new Gestalt.'
\end{axiom}

\begin{axiom}[A9: cross-level input--output isomorphism, homogeneity of levels]\leavevmode
\begin{enumerate}[label=(\alph*)]
\item \textbf{Input--output isomorphism / format-preserving compatibility.} Each level $\ell$ operates over the same representational space $\cU$ (i.e., each level can be understood as an operation mapping $\cU$ to $\cU$), and all levels share the same representational schema (association $G$, similarity sim, confidence degree CD, segmentation/completion predicates). Moreover, because associated representations also belong to $\cU$ (A6), lower-level outputs can become higher-level inputs without conversion, and higher-level Gestalts can be fed back to lower levels without conversion.

\item \textbf{Homogeneity of levels.} The axiomatic rules governing event--representation correspondence, association/generation, selection/retention (and deletion of non-\emph{a priori} representations), and finite capacity constraints operate identically regardless of level. Thus the same axiomatic dynamics can be repeated at each level, providing the formal basis for fractal-like self-similar organization.
\end{enumerate}
\end{axiom}

\begin{axiom}[A10: interconnection of subsystems via Gestalt exchange]
A SANC($E_3$)-intelligent system may be composed of multiple SANC($E_3$)-intelligent subsystems $\{S_i\}$. For any pair $(i, j)$, there exists a communication relation $\mathrm{Comm}(i, j)$ and transfer/reception operators $\mathrm{send}_{i \to j}$, $\mathrm{recv}_{j \leftarrow i}$ that exchange Gestalts (associated events/representations) as units/sets/sequences. Received Gestalts are admitted into the recipient's candidate set $S_t$ and can participate in association (A5--A6), splitting (A7), stabilization and deletion (A3), and segmentation (A8), thereby mutually shaping each subsystem's dynamics. Because a subsystem can itself be recursively organized as a network of subsystems, this axiom supports fractal-like self-similar organization and formalizes Hawkins' ``thousand brains'' hypothesis~\cite{hawkins2021thousand}, extending naturally to multi-agent interaction (Minsky~\cite{minsky1986society}).
\end{axiom}

\section{Derived Propositions}

Each proposition states a consequence generally expected in systems satisfying the axioms. Except T1, T2, T10, the `proofs' for each item are sketches that reveal dependencies on axioms and definitions rather than formal proofs.

\begin{proposition}[T1: no privileged invariant tokens, except system tokens]
Any token other than $\Ssys$ is not exempt from competition, revision, or extinction.

\emph{Dependencies:} A1, A2, A3.

\emph{Proof.} Let $u \in \cU \setminus \Ssys$ be arbitrary. By Axiom A2, $u$ is subject to similarity-based competition: whenever there exists $v \in [u]_{\mathrm{sim}} \cap S_t$ with $v \ne u$, at most one of $\{u, v\}$ can be selected. By Axiom A3(a), non-selection causes CD to decrease. By A3(b)--(c), only \emph{a priori} representations are exempt from forgetting and deletion. Since $u \notin \Ssys$, $u$ receives no exemption: it is subject to competition under A2 and to potential deletion under A3(b). Therefore, $u$ is not exempt from competition, revision, or extinction. \qed
\end{proposition}

\begin{proposition}[T2: inevitable turnover under finite capacity]
For any SANC($E_3$)-intelligent system continuously engaged with an environment, turnover (forgetting) of some Gestalt candidates is unavoidable.

\emph{Dependencies:} A1, A2, A3, A5.

\emph{Proof.} A system continuously engaged with an environment receives events over unbounded time. By A5, co-occurring events generate association candidates, so infinitely many distinct candidates enter $S_t$ over time. By A1, the active capacity is finite at every $t$. By A2, only a limited number can be selected per time step, so some candidates are not selected infinitely often. For any such candidate $u$, A3(a) yields $\CDt(u) \to 0$. By D6, once $\CDt(u) < \theta_{\mathrm{stab}}$, $u \notin \St$, hence $u$ eventually leaves the stabilized set. Therefore turnover is unavoidable. \qed
\end{proposition}

\begin{proposition}[T3: all tokens except system tokens are generated]
Tokens (excluding system tokens) are generated as stabilized structures preferred under $\Esub$ minimization. Only system tokens are structural premises, not products of competition.

\emph{Dependencies:} A3, A4, D1, D6.

\emph{Proof sketch.} A4 drives $\Esub$ minimization. Reducing $\Lrec$ requires representational structures that successfully reconstruct events. A3 reinforces structures that succeed, raising their CD above $\theta_{\mathrm{stab}}$ and thereby conferring token status (D6). Tokens are thus interpretable as emergent outcomes of $\Esub$ dynamics. \qed
\end{proposition}

\begin{proposition}[T4: token--sequence balance]
Minimizing only $\Lrec$ leads to token explosion; minimizing only $\Cstruct$ leads to sequence explosion; $\Esub$ induces a balance between the two explosions.

\emph{Dependencies:} A4.

\emph{Proof sketch.} Direct consequence of A4(c). \qed
\end{proposition}

\begin{proposition}[T5: category formation as compression]
Shared parts or contexts across multiple Gestalts stabilize as new representations (categories), because such categories enable reuse and lower $\Cstruct$ while also reducing $\Lrec$, thereby lowering $\Esub$ doubly.

\emph{Dependencies:} A4, A5, A7, D7.

\emph{Proof sketch.} A7 identifies shared parts across similar Gestalts. Reifying the shared part as a token enables reuse in reconstructing multiple events, lowering $\Cstruct$. A4 therefore favors category formation. \qed
\end{proposition}

\begin{proposition}[T6: prediction is completion]
Given representations that only partially match a Gestalt, $\Esub$ minimization drives completion of the missing parts of the Gestalt; this completion constitutes prediction.

\emph{Dependencies:} A4, D5, D8(a), D8(b).

\emph{Proof sketch.} Partial-match retrieval (D8(a)) retrieves candidate Gestalts sharing parts with current evidence. Replay (D8(b)) internally regenerates them, activating as-yet-unobserved components. This activation is prediction. The derivation formalizes Wertheimer's~\cite{wertheimer1923laws} Gestalt-completion principle. \qed

\emph{Remark (causal inference is a consequence).} ``Causal inference'' is not a separate faculty but a corollary of T6. When events form a temporal Gestalt (e.g., [striking match, flame]), presenting the earlier part activates completion of the unobserved later part, completing the Gestalt---this is experienced as causal inference.
\end{proposition}

\begin{proposition}[T7: hierarchy emerges from closure]
If a single level cannot sufficiently minimize $\Esub$, closure under association (A6) permits recursive composition; $\Esub$ preference (A4) drives formation of multi-level hierarchical structure; subsystem interconnection (A10) allows different subsystems to specialize at different hierarchical levels while sharing representations.

\emph{Dependencies:} A4, A6, A10.

\emph{Proof sketch.} A6 entails that associated structures re-enter $\cU$, enabling recursive processing. Reconstructing complex events at a single level incurs high $\Cstruct$ or high $\Lrec$. Forming higher-level tokens reduces both, so A4 favors hierarchical multi-level structure. A6 and A10 enable horizontal specialization and vertical serialization across subsystems. \qed
\end{proposition}

\begin{proposition}[T8: self-similar, homogeneous levels]
Systems satisfying A1--A10 can form recursively connected multi-level structures, where each level implements dynamics satisfying A1--A10. Thus the overall organization has a self-similar (fractal-like) structure with homogeneity across levels.

\emph{Dependencies:} A6, A9, A10.

\emph{Proof sketch.} By A6, higher-level inputs can be composed from lower-level outputs; by A10, interconnection across levels is possible. Moreover, by A9, the axiomatic role structure repeats in the same form regardless of level, so each level can implement the same axiomatic dynamics (A1--A10). Recursive repetition yields a self-similar structure. \qed
\end{proposition}

\begin{proposition}[T9: system tokens are re-interpretable]
Anchor tokens such as \{here, now, I\} are presupposed as a kind of `relational anchor' that mediates relationships among all tokens. However, their meaning is determined relationally, and as the surrounding tokens and their distances change, the meaning of anchor tokens also evolves.

\emph{Dependencies:} D1, D9, T3.

\emph{Proof sketch.} D1 presupposes the existence of system tokens. D9 defines meaning relationally (via distances to other tokens). By T3, surrounding tokens change and evolve. Hence the relational meaning of system tokens also changes and evolves. \qed
\end{proposition}

\begin{proposition}[T10: unsupervised learning is a consequence]
Internal $\Esub$ minimization under A1--A4 causes learning to occur without explicit external supervision.

\emph{Dependencies:} A1, A3, A4, D9.

\emph{Proof.} By A4(b), the system minimizes $\Esub = \lambda_1 \Lrec + \lambda_2 \Cstruct + \lambda_3 \Cupdate$. By A4(d), reconstruction success strengthens the participating gestalt(s) (increasing CD), while failure triggers updates of $\St$ via segmentation/recombination mechanisms. By A3(a), reinforced representations become more stable, while non-reinforced ones decay. By D9(b), learning is precisely the creation, stabilization, modification, or removal of representations and their relations. Since these state changes are driven solely by internal $\Esub$-minimization and reconstruction-error thresholds (and do not require labels/rewards), learning occurs without external supervision. \qed
\end{proposition}

\begin{proposition}[T11: concepts generated from events form a network: SANC]
The set of stabilized tokens $\St$ forms a \textbf{graph structure} connected by association (e.g., $\costar$ and association operator $G$); in this paper, we call this a `concept network.' Such a network `self-organizes' in the sense that its structure changes and reorganizes without external supervision as tokens are created, deleted, and reinforced according to experience. Anchor tokens serve as globally referenceable reference nodes throughout the system, providing global coordinates (relational anchors) for the network.

\emph{Dependencies:} A5, A6, D1, D4, D6.

\emph{Proof sketch.} A5 forms associations between co-occurring representations. D4 extends association relationships to closure/structure via operator $G$. By D6, the stabilized token set $\St$ is formed, and the connections among tokens and association relationships constitute a graph (network). Anchor tokens from D1 serve as globally referenceable reference points, providing relational coordinates. \qed
\end{proposition}

\begin{proposition}[T12: threshold tuning via system pressure]
$\Esub$ optimization biases the choice of thresholds such as $\theta_{\mathrm{stab}}$. (Too low yields token explosion ($\Cstruct \uparrow$); too high prevents token formation ($\Lrec \uparrow$).) Thresholds are therefore systemically determined near the $\Esub$ optimum.

\emph{Dependencies:} A4, T4, D6.

\emph{Proof sketch.} T4 establishes that $\Esub$ balances token count against sequence complexity. The same trade-off applies to $\theta_{\mathrm{stab}}$: a suboptimal threshold shifts the balance unfavorably. \qed
\end{proposition}

\begin{corollary}[dialogue, reasoning, and related processes]
Under Definition D8(c) (pseudo-memory-mapped I/O) and Axiom A4(d), external sensory input, internal replay (imagination), and action-related signals are processed through the same logical update pathway. Consequently, dialogue, reasoning, and embodied interaction arise as different operational modes of a single representational completion mechanism.

\emph{Justification---formal argument, not a full proof.} By Definition D8(c), externally perceived events and internally replayed representations enter the system through an identical logical data path, differing only in origin, not in processing structure. By Axiom A4(d), each such event---regardless of origin---triggers reconstruction evaluation, coherence reinforcement, or structural update of $\St$ according to reconstruction success or failure. Hence, no formal distinction is made within the update dynamics between perception-driven input and internally generated input. Action-related signals, insofar as they are evaluated by their reconstructive consequences on subsequent sensory input, are integrated into the same update loop. Therefore, dialogue (linguistic interaction), reasoning (internal replay and completion), and physical interaction (action--perception cycles) are unified as instances of representational completion under a common update mechanism. \qed

\emph{Remark (\textbf{embodiment and physical AI}).} The corollary extends naturally to embodied agents. Motor action can be understood as the external realization of an internally completed Gestalt: the system replays a Gestalt whose completion includes motor-effector components, and this replay causes physical changes in the environment. The resulting environmental change produces new sensory input, which re-enters the same axiomatic pathway (D8(c)). Thus the perception--action loop is not a separate control module but a continuous chain of Gestalt completions under $\Esub$ minimization. Physical AI, robotics, and embodied cognition require no additional axioms; they are automatic consequences of the framework.
\end{corollary}

\section{Extended Definitions: Higher Cognition as Gestalt Completion}

\begin{definition}[D10: desire/drive]
Desire is a bias toward internally completing and confirming a selected Gestalt.
\end{definition}

\begin{definition}[D11: will/intent; selection bias]
Will is a bias that disposes the system to select a particular Gestalt among multiple candidates---a bias applied to the selection process itself. Assume that at time $t$ each candidate $u \in S_t$ has a scalar selection score $\mathrm{score}_t(u) \in \mathbb{R}$, and that will introduces an additive bias $\Delta_t^{\mathrm{will}}(u) \in \mathbb{R}$. The selection score is defined as:
\[
\mathrm{score}_t^{\mathrm{will}}(u) := \mathrm{score}_t(u) + \Delta_t^{\mathrm{will}}(u).
\]
Define a score-based selection operator (e.g., argmax or top-$k$), and define will-biased selection as:
\[
\Pi_t^{\mathrm{will}}(S_t) := \arg\max_{u \in S_t} \mathrm{score}_t^{\mathrm{will}}(u) \quad \text{(or top-}k\text{ for } k > 1\text{)}.
\]

\emph{Remark.} This may be connected to Schopenhauer's notion of ``will''~\cite{schopenhauer1819world}, but this is an interpretive gloss, not part of the axioms.
\end{definition}

\begin{definition}[D12: self]
Let the system token ``I'' $\in \Sanchor$ function as an anchor for first-person self-reference. At time $t$, define the set of tokens proximate to ``I'' as:
\[
X_{I,t} := \{ \tau \in \St : d(\tau, I) \le \delta_I \}.
\]
For any set $X \subseteq \cU$, define the association-closure under $G$ $\mathrm{cl}_G(X) \subseteq \cU$ as ``the smallest set containing $X$ and closed under association operator $G$ (A6).'' The self-representation $u_{\mathrm{self}}$ is a representative Gestalt selected (and reinforced) within this closure:
\[
u_{\mathrm{self}} \in \Pt(\mathrm{cl}_G(X_{I,t})).
\]
Thus the self is a Gestalt composed from tokens proximate to the anchor ``I'' and strengthened through selection; the specific association order or implementation is not prescribed.

\emph{Remark.} Because this construction forms a closure containing the anchor token ``I,'' it suggests the possibility of interpreting phenomena related to Hofstadter's ``strange loop''~\cite{hofstadter1979geb} and self-referential representations in logic and set theory. This remark is interpretive, not part of the axioms.
\end{definition}

\begin{definition}[D13: thought, emotion, sleep]\leavevmode
\begin{enumerate}[label=(\alph*)]
\item \textbf{Thought:} The process of retrieving higher-level Gestalts consistent with lower-level Gestalts, replaying subsequent Gestalts, and having the self treat them again as events to be represented.
\item \textbf{Emotion:} A Gestalt containing evaluative information (whether beneficial or detrimental) relevant to maintaining the existence of the self, together with associated sensory-source tokens.
\item \textbf{Sleep:} The process of testing whether (temporarily) formed Gestalt candidates can replay to reconstruct experience while reducing or blocking external input. Replay failure triggers reconstruction or new formation to reduce $\Esub$ and improve restoration performance.
\end{enumerate}
\end{definition}

\begin{definition}[D14: dialogue, authoring, perception, social interaction]\leavevmode
\begin{enumerate}[label=(\alph*)]
\item \textbf{Dialogue:} Given a question Gestalt $Q$, retrieving stabilized question--answer $(Q, A)$ Gestalts and activating the answer Gestalt $A$ to complete the QA Gestalt, or generating a new $(Q, A)$ Gestalt.
\item \textbf{Authoring:} Under partial cues, desire and will (D11) select a compatible higher-level Gestalt; subsequent Gestalts are internally replayed and cause corresponding external events (realizing the replayed representations as scores, sentences, or drawings). Authoring is complete when the system verifies that the selected Gestalt completes consistently with the intended bias (D11).
\item \textbf{Perception (vision/speech):} Partial sensory streams activate candidate Gestalts, which compete to complete latent structures that explain the input. The percept is the operation of internally finding a stabilized Gestalt that best satisfies finite capacity and $\Esub$, and mapping the given events to the Gestalt.
\item \textbf{Social interaction as a consequence:} Multi-agent interaction requires no additional axioms. Other agents are external event sources to an individual intelligence (agent); D14 formalizes turn-taking as question--answer completion; A10 models theory of mind via interconnected subsystems---a formalization of Minsky's ``society of mind''~\cite{minsky1986society}.
\end{enumerate}
\end{definition}

\begin{definition}[D15: self-improvement]
`Self-improvement' refers to the process in which a `self (D12)' representation is generated within the system, emotion (D13) induces will (D11) in a direction favorable to maintaining the self's existence, biasing/guiding the progression of thought (D13), and as a result, a new self is authored (D14).
\end{definition}

\section{Discussion -- Relationship to Existing Theories}

\begin{table}[ht]
\centering
\small
\begin{tabular}{@{}lll@{}}
\toprule
\textbf{Theory} & \textbf{Shared Insight} & \textbf{SANC Distinction} \\
\midrule
Global Workspace~\cite{baars1988cognitive} & Competition for access & Similarity-based, not broadcast-based \\
IIT~\cite{tononi2004information} & Integration under constraints & Binding-based, not $\Phi$-maximizing \\
Predictive processing~\cite{clark2013whatever} & Prediction at the core & Prediction = completion (T6) \\
Free Energy Principle~\cite{friston2010free} & Energy-based dynamics & $\Esub$ explicitly decomposed \\
Transformers/LLMs~\cite{vaswani2017attention} & Token-based prediction & Tokens are outcomes (T3) \\
Forward--Forward~\cite{hinton2022forward} & Local learning & Formalized via A2, A3 \\
Nested Learning~\cite{behrouz2025nested} & Multiple time scales & Formalized via A3, T8 \\
Hofstadter (GEB)~\cite{hofstadter1979geb} & Strange loops & Derived from A6, A9, D12 \\
Minsky (Society of Mind)~\cite{minsky1986society} & Interacting agents & Formalized via A10 \\
Kantian \emph{a priori}~\cite{kant1781kritik} & Structural premises & System tokens as network nodes \\
AIXI~\cite{hutter2005universal} & Formal optimal agent & Computable, representation-emergent \\
NARS~\cite{wang2019defining} & Finite resources, adaptation & Adds token emergence, hierarchy, energy \\
Chollet~\cite{chollet2019measure} & Generalization efficiency & Generative constraints, not measurement \\
\bottomrule
\end{tabular}
\caption{Comparison of SANC($E_3$) with existing theories.}
\label{tab:comparison}
\end{table}

\section{Conclusion}

SANC($E_3$) formalizes general intelligence as an axiomatic framework of an event--representation system oriented toward $\Esub$ minimization under finite capacity. This paper has established an axiomatic characterization of general intelligence. Ten axioms---five foundational (A1--A5) and five structural (A6--A10)---together with the energy functional $\Esub$ yield twelve derived propositions. These propositions demonstrate that token formation, forgetting, category emergence, hierarchy, and self-similar organization are not independent design choices but mutually entailed structural necessities. Three results merit emphasis:

\textbf{Tokens as outcomes.} Proposition T3 establishes that all tokens except system tokens are products of $\Esub$ dynamics. This inverts the standard assumption: vocabulary is not an input to intelligence but an output of it.

\textbf{Cognition as completion.} Proposition T6 and the Corollary unify prediction, dialogue, authoring, perception, causal inference, and embodied action as instances of Gestalt completion. No separate modules are required; a single mechanism suffices.

\textbf{Inevitability of forgetting.} Proposition T2 proves that any system satisfying the axioms must forget. Memory is not merely storage but active curation under finite capacity---a feature, not a limitation.

\subsection{Implications}

If these axioms capture necessary conditions for general intelligence, then scaling alone cannot bridge the gap between current systems and AGI. What is missing is not more parameters but the right structural constraints: autonomous token formation (T3), similarity-based competition (A2), and the reconstruction--compression--stability trade-off (A4). Conversely, any system satisfying A1--A10 would exhibit the derived properties by construction. The question ``What is intelligence?'' admits no single answer. SANC($E_3$) offers one: intelligence is $\Esub$-minimizing Gestalt completion under finite capacity, anchored by system tokens, generating its own representational vocabulary. Whether this characterization is correct is ultimately an empirical question---one we intend to pursue.

\subsection{Future Work}

This axiomatic system specifies what must exist in a system called intelligent but does not prescribe a unique algorithm that satisfies it. Many `near-intelligent' systems exist, from attention-based Transformers~\cite{vaswani2017attention} to GPT~\cite{radford2018improving,brown2020language}, but human intelligence remains the only system satisfying all these axioms. Implementing an algorithm that fully satisfies this axiomatic system is a necessary step toward achieving artificial general intelligence (AGI).

We are developing a prototype that induces tokens from raw character streams across three hierarchical levels. Demonstrating that such a system satisfies A1--A10 and manifests T1--T12 would provide empirical grounding. Extending the framework to embodied agents---coupling $\Esub$ to irreversible motor consequences---remains the principal theoretical task.

\subsection{Acknowledgements}

This work stands on a long intellectual lineage of humanity. From Confucius's insight~\cite{confucius500bce} that ``to know what you know and to know what you do not know---that is knowledge,'' to Plato's epistemology~\cite{plato369bce} that ``knowledge is justified true belief''; from the debates of Yi Hwang~\cite{yihwang1568} and Yi I~\cite{yii1575} on the essence and priority of the Four Beginnings and Seven Emotions, to Kant~\cite{kant1781kritik} who viewed space and time as \emph{a priori} forms of experience, and Schopenhauer~\cite{schopenhauer1819world} who grasped the world as will and representation.

From Wertheimer's~\cite{wertheimer1923laws} Gestalt psychology---``the whole is more than the sum of its parts''---to Saussure's~\cite{saussure1916cours} structural linguistics in which linguistic units are defined relationally; from Hebb's~\cite{hebb1949organization} principle that ``neurons that fire together wire together,'' to Quillian~\cite{quillian1968semantic} who proposed associative networks of concepts.

From Hofstadter's~\cite{hofstadter1979geb} ``strange loops'' and ``self-reference,'' to Minsky's~\cite{minsky1986society} ``society of mind,'' to Penrose~\cite{penrose1989emperor} who raised the non-algorithmic nature of consciousness---for over 2,000 years, all have asked the same question: What is knowledge? \textbf{What is intelligence?}

To this question---perhaps the oldest since humanity's emergence---SANC($E_3$) attempts to offer one answer. By stating in axiomatic form what any generally intelligent system must satisfy, and by integrating the insights of our predecessors into a minimal formal framework. We hope this attempt proves meaningful to all who have not stopped asking this question throughout humanity's long existence, and to all who will continue to ask it.

\bibliographystyle{plain}
\bibliography{main}

\appendix

\section{Summary of Token Types, Axioms, and Propositions}
\label{app:summary}

\textbf{Token types}
\begin{itemize}
\item System tokens ($\Ssys$): Structurally given, including anchor tokens \{here, now, I\} and sensory-source tokens.
\item Tokens ($\St \setminus \Ssys$): Representational units that emerge via self-organization and $\Esub$ dynamics.
\end{itemize}

\textbf{Axioms}

\begin{table}[ht]
\centering
\begin{tabular}{@{}ll@{}}
\toprule
\textbf{Axiom} & \textbf{Description} \\
\midrule
A1 & Finite capacity \\
A2 & Similarity-based competition \\
A3 & CD stabilization \\
A4 & Energy $\Esub$ \\
A5 & Co-occurrence $\to$ association candidates \\
A6 & Closure \\
A7 & Splitting \\
A8 & Uncertainty-gated segmentation \& Gestalt completion \\
A9 & Isomorphism across levels \\
A10 & Subsystem interconnection \\
\bottomrule
\end{tabular}
\end{table}

\textbf{Propositions}

\begin{table}[ht]
\centering
\begin{tabular}{@{}lll@{}}
\toprule
\textbf{Thm} & \textbf{Description} & \textbf{Dependencies} \\
\midrule
T1 & No privileged tokens (except $\Ssys$) & A1, A2, A3 \\
T2 & Memory requires forgetting & A1, A2, A3, A5 \\
T3 & Tokens are outcomes & A3, A4, D1, D6 \\
T4 & Token--sequence balance & A4 \\
T5 & Categories as compression & A4, A5, A7, D7 \\
T6 & Completion = prediction & A4, D5, D8(a), D8(b) \\
T7 & Hierarchy from closure & A4, A6, A10 \\
T8 & Self-similar levels & A6, A9, A10 \\
T9 & System tokens re-interpretable & D1, D9, T3 \\
T10 & Unsupervised learning & A1, A3, A4, D9 \\
T11 & Concept network & A5, A6, D1, D4, D6 \\
T12 & Threshold tuning & A4, T4, D6 \\
\bottomrule
\end{tabular}
\end{table}

\section{Mapping from Original SANC Postulates}
\label{app:mapping}

The SANC($E_3$) formalization is based on the authors' original SANC postulates (C.1--C.31). The table below maps the original postulates to definitions, axioms, and propositions in this paper.

\begin{longtable}{@{}ll@{}}
\toprule
\textbf{Original Postulate} & \textbf{This Paper} \\
\midrule
\endhead
C.1 (events $\to$ representations) & id in D1 \\
C.2 (distinction and consistency) & D2 \\
C.3 (co-occurrence and association/Gestalt) & A5, D4, D5 \\
C.4 (similarity and splitting) & D3, A7 \\
C.5 (category generation) & D7, T5 \\
C.6 (partial-match retrieval) & D8(a) \\
C.7 (internal replay/regeneration) & D8(b) \\
C.8 (forgetting) & A3(b) \\
C.9 (restoration and active update) & A4(d) \\
C.10 (completion candidacy via uncertainty) & A8 \\
C.11 (reinforcement/exclusion) & A3, D11 \\
C.12 (competition and survival) & A2 \\
C.13 (capacity constraint: token vs.\ sequence explosion) & A1, T4 \\
C.14 (automatic threshold determination) & T12 \\
C.15 (multi-level identical structure) & T8 \\
C.16 (I/O isomorphism across levels) & A9 \\
C.17 (concept network: SANC) & T11 \\
C.18 (interconnected subsystems) & A10 \\
C.19 (embedding by co-occurrability) & D9 \\
C.20 (\emph{a priori} representations) & D1, $\Ssys$ \\
C.21 (self as embedding) & D12, T8 \\
C.22 (desire and will) & D10, D11 \\
C.23 (thought) & D13 \\
C.24 (emotion) & D13 \\
C.25 (sleep) & D13 \\
C.26 (dialogue) & D14 \\
C.27 (creativity/writing) & D14 \\
C.28 (vision/speech processing) & D14 \\
C.29 (unsupervised learning) & T10 \\
C.30 (self-improvement; environment improvement) & T10, Section~7 \\
C.31 (unintentional intelligence) & Section~7 \\
\bottomrule
\end{longtable}

\section{Original SANC Postulates (Extended)}
\label{app:postulates}

This appendix presents the original SANC postulates, first formulated in the mid-1990s, in informal language. Many readers may find these more intuitive and accessible than the formal axioms in the main text.

\medskip

\textbf{C.1 From events to representations.} A detected event generates a corresponding internal representation. An ``event'' may be sensory, motor, interoceptive, linguistic, abstract, or any detectable form.

\textbf{C.2 Distinction and consistency.} Intelligence can distinguish identical, similar, and different events, and can assign consistent internal representations to them.

\textbf{C.3 Co-occurrence and association (Gestalt).} Events detected together become associated. This associated event is a ``Gestalt.'' \emph{Corollary:} A Gestalt can be treated as an event at a higher level and can be recombined.

\textbf{C.4 Similarity and splitting.} Similar events split into shared and differing parts (e.g., $ab =$ `with mom' and $ac =$ `with dad' share $a =$ with but differ in $b =$ mom vs.\ $c =$ dad).

\textbf{C.5 Category (company) generation.} If the differing parts co-occur with other events, categories are formed both for the shared parts with those other events and for the differing parts themselves. \emph{Example:} If $b$ and $c$ also appear with $d$ (i.e., $db =$ `for mom', $dc =$ `for dad' are experienced), then two categories [$a$, $d$] and [$b$, $c$] are formed (e.g., [with, for] and [mom, dad]).

\textbf{C.6 Partial-match retrieval.} Using stored Gestalts, the system can retrieve Gestalts that partially match a given event sequence.

\textbf{C.7 Internal regeneration and reactivation.} The system retrieves and internally regenerates/reactivates Gestalts that match the event sequence so far. \emph{Corollary:} Internal regeneration fills ``gaps'' in the given sequence.

\textbf{C.8 Forgetting.} Non-\emph{a priori} representations (including those generated from experienced events and \emph{a priori} representations) are deleted if the corresponding event is not re-experienced within a certain time.

\textbf{C.9 Restoration and active updating.} The system restores all events so far using Gestalts (and actively updates Gestalts until restoration becomes complete).

\textbf{C.10 Completion candidacy via uncertainty.} As an experienced sequence grows, predicting the next event becomes easier and more certain; however, when a context is completed and a new context begins, next-event prediction suddenly becomes difficult. In this case, the sequence up to that point is treated as a completed-Gestalt candidate.

\textbf{C.11 Reinforcement and exclusion.} The more often a Gestalt is selected, the stronger it becomes; associated events that frequently lose in competition are excluded (via CD).

\textbf{C.12 Competition and survival.} Candidate Gestalts compete to explain the currently given events, and only one is selected.

\textbf{C.13 Capacity constraint (token explosion vs.\ sequence explosion).} Associated-event creation should be minimized. Creating higher-level Gestalts at every level causes a combinatorial explosion of representational units (token explosion). Failing to create higher-level Gestalts forces storage of too many event representations to restore experience (sequence explosion). With finite resources, intelligence balances restorability of experienced sequences against creation of new higher-level tokens.

\textbf{C.14 Automatic threshold determination.} Thresholds that suppress new higher-level token creation (balancing token vs.\ sequence explosion) and thresholds on CD that make a token persistent are not fixed \emph{a priori}; they are determined dynamically by competition (C.12).

\textbf{C.15 Multi-level identical structure.} Intelligence consists of multiple levels with the same structure (fractal-like).

\textbf{C.16 Input--output isomorphism across levels.} Every level takes inputs of the same form and produces outputs of the same form.

\textbf{C.17 Network of concepts (SANC).} Each level produces higher-level events as associations of input events; such higher-level events form a network (graph) of co-occurring or co-occurrable lower-level events. Viewing events as concepts, intelligence is the system that automatically and actively generates/organizes concepts and their networks in response to events.

\textbf{C.18 Interconnected subsystems.} Intelligence is a system of multiple such multi-level subsystems that are interconnected. Gestalt sets are exchanged among these vertical and horizontal interconnections, enabling formation and processing of different levels of concepts (representations) and different modalities.

\textbf{C.19 Embedding (based on co-occurrability).} Every event is embedded by the events it co-occurs with or can co-occur with, and by distances within that relation network. That is, embedding is the information contained in the `co-occurrence/co-occurrability' relation network.

\textbf{C.20 \emph{A priori} representations exist.} Some representations exist intrinsically rather than being generated from others (e.g., raw sensory signals; character codes in computers). Some \emph{a priori} representations are embedded in all events (e.g., ``now,'' ``here,'' ``I'').

\textbf{C.21 Self.} The self is a representation (embedding) of the intelligence itself, and a set of representations lying within a certain distance from the \emph{a priori} representation ``I.''

\textbf{C.22 Desire and will (selection bias).} Desire is the tendency to internally complete (realize) a Gestalt from given events. Will is a bias that makes certain Gestalts among many selectable candidates be selected/regenerated more often.

\textbf{C.23 Thought.} A process in which a top-level Gestalt retrieves and internally represents/constructs the next Gestalt at the next-lower level.

\textbf{C.24 Emotion.} A Gestalt that encodes information about being favorable or unfavorable for maintaining the existence of the ``self.''

\textbf{C.25 Sleep.} A process that tests whether (temporarily) formed Gestalts can reconstruct experienced reality while reducing/blocking external input. If reconstruction fails, the system reconstructs/creates Gestalts and repeats internal replay to improve restoration performance.

\textbf{C.26 Dialogue.} Treating a question--answer (QA) pair as a unit Gestalt, a dialogue is a sequence of QA Gestalts. When only Q is given from outside, completion pressure selects/regenerates an appropriate A to complete the QA Gestalt; this is answer generation.

\textbf{C.27 Creativity/writing.} Completing a partially specified Gestalt (topic/constraints/structure) via replay/regeneration in a direction that reduces $\Esub$, producing a novel structured artifact.

\textbf{C.28 Visual/speech processing.} When partial sensory streams activate candidate Gestalts, the candidates compete to complete a latent structure that explains the input. Recognition is the stabilized completion that best fits under finite capacity and $\Esub$.

\textbf{C.29 Unsupervised learning.} A system with the above properties can learn from event sequences without explicit external supervision.

\textbf{C.30 Self-improvement and environment improvement.} Such a system can select or create environments advantageous for survival and can improve itself.

\textbf{C.31 Unintentional intelligence.} Functions often called the ``essence'' of intelligence---prediction, memory, learning, classification, creativity, reasoning, etc.---arise necessarily and as by-products in systems that satisfy the SANC postulates.

\end{document}